\documentclass[10pt,twocolumn,letterpaper]{article}
\usepackage[letterpaper,text={6.75in,9in},centering,columnsep=0.25in]{geometry}
\usepackage[T1]{fontenc}
\usepackage{mathptmx}
\usepackage{times}
\usepackage{amsmath,amssymb,bm}
\usepackage{booktabs}
\usepackage{graphicx}
\usepackage{xcolor}
\usepackage{url}
\usepackage[round]{natbib}

\usepackage[font=small,labelfont=bf,skip=4pt]{caption}
\usepackage{titlesec}
\usepackage{microtype}
\usepackage[hidelinks]{hyperref}
\titleformat{\section}{\normalfont\large\bfseries}{\thesection}{0.6em}{}
\titleformat{\subsection}{\normalfont\normalsize\bfseries}{\thesubsection}{0.6em}{}
\titlespacing*{\section}{0pt}{9pt plus 2pt minus 1pt}{4pt plus 1pt}
\titlespacing*{\subsection}{0pt}{7pt plus 2pt minus 1pt}{3pt plus 1pt}
\newcommand{\sysname}{TalkMesh}
\newcommand{\ind}[1]{\mathbf{1}\!\left[#1\right]}
\graphicspath{{figures/}}
\begin{document}
\twocolumn[{%
\begin{center}
{\LARGE\bfseries Reinforcement Learning of Communication in a Mesh of Small Language Models\par}\vspace{10pt}
{\large Mehmet Kerem Turkcan\\[2pt]
{\normalsize Department of Civil Engineering, Columbia University}\\
{\normalsize\texttt{mkt2126@columbia.edu}}\par}
\end{center}
\vspace{4pt}
}]
\begin{center}\textbf{\large Abstract}\end{center}\vspace{-4pt}
{\small Language models gain accuracy from more compute at test time, but majority voting over independent samples saturates: as samples grow, the vote converges to the model's most frequent answer. Communication can add what sampling cannot: an agent that solves a problem can pass the key step to the others. We present \sysname{}, a decentralized mesh of small language model agents that learns when and what to communicate. Each agent samples a proposal and scores it with a trained confidence head. The most confident agent broadcasts a hint; agents below a confidence threshold revise, keeping each revision that outscores its proposal. Gossip consensus approximates the vote weighted by confidence without a coordinator. A talk policy, trained with group relative policy optimization on the change in correctness after revision, writes hints and revisions. With three agents, which together generate at most six outputs, the mesh reaches the accuracy of majority voting over 32 samples with each of three models. Trained with at most 8 agents and evaluated with 32, it raises accuracy from 0.568 under self-consistency to 0.705 (Qwen3.5-0.8B, GSM8K) and from 0.492 to 0.722 (SmolLM3-3B, MATH-500). When 4 of 8 agents collude on a wrong answer with fabricated confidence and poisoned hints, majority vote accuracy falls to 0.000 (Qwen3.5-0.8B, GSM8K). A defended mesh, whose agents rescore solutions with their own confidence heads, retains 0.507. Across reasoning, embodied coordination, and traffic signal control, messages improve a decision when the acting agent cannot observe the information it requires and another agent can send it.\par}

\section{Introduction}
\label{sec:intro}

\begin{figure*}[t]
  \centering
  \includegraphics[width=\textwidth]{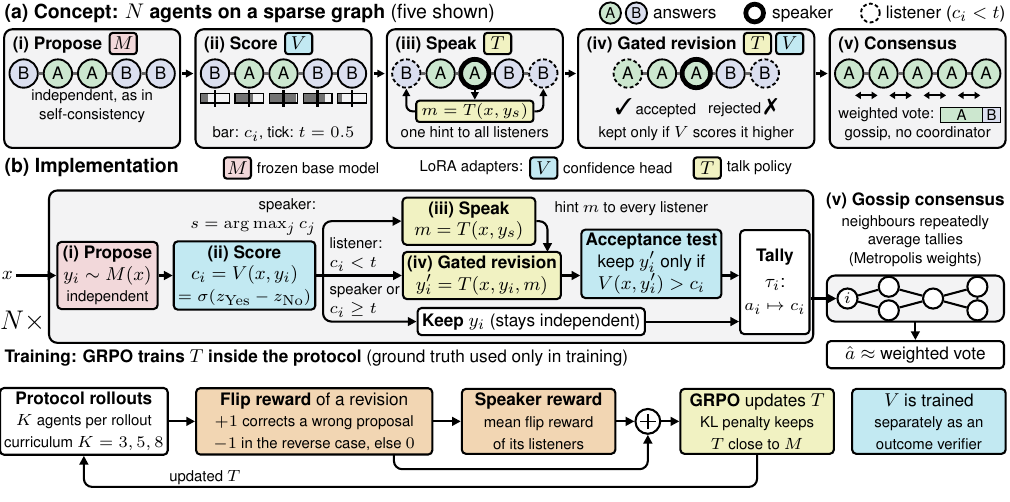}
  \caption{Overview of \sysname. (a) Concept: independent proposals, one hint from the most confident agent, revisions gated by the confidence head, and gossip consensus. (b) Implementation: the frozen base model with two adapters, the five inference stages, and the training loop of the talk policy.}
  \label{fig:overview}
\end{figure*}

Self-consistency \citep{wang2023selfconsistency} scales the compute of a language model at test time: it samples independent chains of thought and returns the majority answer. Accuracy increases with the number of samples and saturates as the vote converges to the model's most frequent answer. Reading a peer's reasoning lets an agent correct errors and recover omitted steps, so interaction adds information. In multiagent debate and discussion \citep{du2023debate,liang2023encouraging}, a fixed set of agents reads one another's solutions in rounds designed by hand. Agents that converge on the same answers correlate their errors, which removes the independence the majority vote relies on. Inference with interacting agents therefore requires communication that adds information and preserves independence.

We present \sysname{}, a decentralized mesh of small language model agents on a sparse communication graph, whose agents learn when and what to communicate (Figure~\ref{fig:overview}). Each agent first samples a proposal $y_i$ from the frozen base model $M$, as in self-consistency. We attach two LoRA adapters to $M$: the confidence head $V$ scores every proposal, and the talk policy $T$ writes hints and revisions. The agent with the highest score, the speaker, writes a hint. Every listener, an agent scoring below a threshold, reads the hint and rewrites its proposal as a revision $y_i'$. All other agents keep their proposals, which remain independent. The acceptance test replaces a proposal with its revision only if $V$ scores the revision higher. Gossip consensus, in which neighbours repeatedly average answer tallies, approximates the vote weighted by confidence scores (the weighted vote) without a coordinator. We train $T$ inside the protocol with group relative policy optimization (GRPO) on a flip reward, $+1$ for a revision that corrects a wrong proposal and $-1$ for the reverse; a hint receives the mean flip reward of its listeners. A curriculum over network size raises the rollout size $K$ (agents per training rollout) through 3, 5, and 8.

Figure~\ref{fig:scaling} reports accuracy against network size $N$. With three agents, the trained mesh reaches the accuracy of self-consistency with 32 samples for all three models. At $N=32$ (200 test problems per model), the trained mesh raises accuracy from 0.492 under self-consistency to 0.722 for SmolLM3-3B on MATH-500 and from 0.568 to 0.705 for Qwen3.5-0.8B on GSM8K; without messages, the weighted vote (Rerank) reaches 0.590 and 0.655 (Table~\ref{tab:instrumental}). For these two models and Qwen3-1.7B on MATH-500, the trained mesh exceeds self-consistency at every $N$ from 3 to 32 after training with at most $K=8$ agents. When 4 of $N{=}8$ agents are compromised, submitting a shared wrong answer with claimed confidence 0.99 and a poisoned hint as speaker, majority vote accuracy on 150 test problems falls to 0.000 for all three models. The defended mesh rescores every solution with the confidence head and applies the acceptance test to every revision. It retains 0.507 with Qwen3.5-0.8B on GSM8K (0.667 without attack) and exceeds the undefended mesh, which omits both checks and falls to at most 0.007, for every model ($p<10^{-3}$, McNemar's exact test; Section~\ref{sec:attack}). On the embodied coordination benchmark SwarmBench, REINFORCE training raises the Flocking score of all three models over twenty unseen seeds, and a Qwen3.5-0.8B policy distilled from a centralized solver matches the solver on Synchronization (Section~\ref{sec:swarm}). We introduce a traffic signal control benchmark of sixteen intersections under New York City timing constraints and train language model meshes by behaviour cloning on traces of a communicating controller whose incident warnings reach intersections that cannot observe the incident. Before any language model evaluation, we fix a severe subset of ten of the 70 scenarios, on which messages reduce the communicating controller's mean person delay by 48.2\,s. On this subset, delivering messages reduces the person delay of the Qwen3.5-0.8B, 2B, and 4B meshes by $23.6 \pm 6.7$\,s (standard error over 60 paired episodes). Over all 70 scenarios, the reduction for the 4B mesh is $11.4 \pm 6.9$\,s (Section~\ref{sec:traffic}).

The hint and the incident warning illustrate the observability criterion: communication improves a decision when the required information is unobservable to the acting agent and observable to another agent that can send it. Withholding messages lowers performance in reasoning and traffic, where the acting agent cannot observe the required information; in two SwarmBench settings where it can, scores with and without messages agree within one standard error (Table~\ref{tab:instrumental}).
\section{Related work}
\label{sec:related}

Self-consistency \citep{wang2023selfconsistency} established majority voting over sampled chains of thought as the baseline for inference scaling. \citet{cobbe2021verifiers} showed that reranking by a trained verifier outperforms finetuning alone, and \citet{li2024moreagents} showed that voting accuracy increases with the number of agents across models and tasks. Without communication, the mesh reduces to the weighted vote (Rerank), which combines voting with verifier reranking. Multiagent debate \citep{du2023debate,liang2023encouraging}, ReConcile \citep{chen2024reconcile}, and Exchange-of-Thought \citep{yin2023eot} use handcrafted discussion rounds and an untrained communication policy among a fixed number of agents. Decentralized agent networks \citep{yang2025agentnet,li2025swarmsys,ruan2025swarmbench,zhang2026silobench,tian2026queenbee} and consensus schemes robust to Byzantine agents \citep{jo2025byzantine,lee2026byzantine} use frozen models. Our training is closest to reinforcement learning of agent collaboration \citep{liu2025magrpo,sun2024llmbased}, learned multiround message protocols \citep{zhang2025learning,fan2026todycomm}, communication guided by language models for cooperative control \citep{bae2026llmguided}, and emergent communication in multiagent reinforcement learning \citep{foerster2016learning,sukhbaatar2016learning,lowe2017maddpg}. In \sysname{}, (i) messages are natural language, (ii) the confidence head selects speakers and listeners, (iii) each hint receives the mean flip reward of its listeners, and (iv) we train $T$ inside the inference protocol.

Concurrent work with frontier models shows that agents sharing verified progress through a common workspace outperform the best independent agent \citep{park2026testtime}. Teams with collaboration strategies learned by reflective prompt search exceed a perfect router over their members' answers in average accuracy on mathematics and physics benchmarks \citep{pappu2026selforganizing}. Both find that communication helps most when progress is verifiable; our acceptance test verifies each revision with the confidence head. \sysname{} trains the talk policy with GRPO on at most 8 agents, reaches consensus by gossip without a judge model or common workspace, and exceeds self-consistency at every $N$ up to 32 with five open models of 0.8B to 4B parameters (Sections~\ref{sec:scaling} and~\ref{sec:family}).
\section{Method}
\label{sec:method}

Each agent in \sysname{} runs a frozen base model $M$ with two LoRA adapters: (i) a confidence head $V$, which scores a solution, and (ii) a talk policy $T$, which writes hints and revisions (Figure~\ref{fig:overview}). Agents with the same base model share these weights; in a heterogeneous mesh (Section~\ref{sec:attack}), each agent scores and revises with the adapters of its own model. Because $M$ is never updated, training leaves the accuracy of a single proposal unchanged.

\textbf{Inference protocol.} Given a problem $x$ and $N$ agents on a connected communication graph, the mesh executes five stages. (i) \emph{Propose}: each agent samples a proposal $y_i \sim M(x)$ with both adapters disabled and the sampler of self-consistency, so the extracted answers are i.i.d. Answers come from the \texttt{ANSWER:} tag (fallback: a boxed answer) and match numerically; a proposal without an answer abstains. (ii) \emph{Score}: the confidence head assigns $c_i = V(x, y_i) = \sigma(z_{\text{Yes}} - z_{\text{No}})$, the sigmoid of the logit gap between \emph{Yes} and \emph{No} after a fixed verification prompt. (iii) \emph{Speak}: the speaker is the agent with the highest confidence score, $s = \arg\max_i c_i$. Agents find $s$ without a coordinator by gossiping the maximum $c_i$ within a number of rounds equal to the graph diameter. The speaker generates a hint $m = T(x, y_s)$ of at most 112 tokens; the speaker prompt requests the key idea, the critical step, and the final answer. (iv) \emph{Gated revision}: every listener (agent $i \neq s$ with $c_i$ below the threshold $t = 0.5$ or without an answer) generates a revision $y_i' = T(x, y_i, m)$. The acceptance test keeps the revision only if $V(x, y_i') > V(x, y_i)$. The other agents keep their proposals, which remain independent. (v) \emph{Consensus}: gossip consensus computes the mesh answer as the weighted vote $\hat{a} = \arg\max_a \sum_i c_i \,\ind{a_i = a}$, with answers $a_i$ and scores $c_i$ after stage (iv).

\textbf{Gossip consensus.} Each agent $i$ holds a sparse tally vector $\tau_i$ over answer strings, initialized as $\{a_i \mapsto c_i\}$, and in each gossip round $r$ averages it with the tallies of its neighbours:
\begin{equation}
\label{eq:gossip}
\begin{gathered}
\tau_i^{(r+1)} = W_{ii}\,\tau_i^{(r)} + \sum_{j \in \mathcal{N}(i)} W_{ij}\,\tau_j^{(r)},\\[2pt]
W_{ij} = \frac{1}{1 + \max(d_i, d_j)},\qquad W_{ii} = 1 - \sum_{j \in \mathcal{N}(i)} W_{ij},
\end{gathered}
\end{equation}
where agent $i$ has neighbours $\mathcal{N}(i)$ and degree $d_i$. These Metropolis weights form a symmetric, doubly stochastic mixing matrix, so on any connected graph every $\tau_i^{(r)}$ converges to the network average \citep{xiao2005scheme,boyd2006gossip}. The answer of each agent, $\arg\max_a \tau_i^{(r)}(a)$, therefore converges to the centralized weighted vote; Section~\ref{sec:scaling} measures their agreement. Over $R$ gossip rounds, agent $i$ sends $O(d_i R)$ tallies, independent of $N$; each tally holds one entry per distinct answer. All runs use connected Watts-Strogatz graphs with rewiring probability $0.1$, mean degree $4$ ($2$ at $N{=}3$), and $R = 25$.

\textbf{Training the confidence head.} For each model and dataset, we train $V$ as an outcome verifier \citep{cobbe2021verifiers} on eight solutions sampled from $M$ per training problem, labelled by the ground truth answer. Batches balance correct and incorrect solutions, and we keep the checkpoint with the highest validation accuracy.

\textbf{Training the talk policy.} We train $T$, one adapter for speaker and listener initialized to reproduce $M$, with group relative policy optimization \citep[GRPO;][]{shao2024deepseekmath} on rollouts of the inference protocol. A rollout runs stages (i) to (iii) and selects the listeners. For each problem, we sample a group of $G{=}4$ hints $m_1, \dots, m_G$ from the speaker context $(x, y_s)$, and every listener writes one revision under each hint. The revision $y_{g,i}'$ of listener $i$ under hint $g$ receives the flip reward
\begin{equation}
r_{g,i} = \ind{y_{g,i}' \text{ correct}} - \ind{y_i \text{ correct}} \in \{-1, 0, +1\},
\end{equation}
and hint $g$ receives the speaker reward $R_g = \frac{1}{|L|}\sum_{i \in L} r_{g,i}$, the mean flip reward over listeners $L$. Both rewards score revisions before the acceptance test. We obtain advantages by standardizing (i) speaker rewards over the $G$ hints of one problem and (ii) flip rewards over the $G$ revisions of one listener. Problems without listeners and groups with equal rewards contribute no gradient. We penalize the KL divergence of $T$ from $M$ with the $k_3$ estimator (weight $0.1$) \citep{schulman2020kl}. Only training uses ground truth answers, as in centralized training with decentralized execution \citep{lowe2017maddpg}. A curriculum over network size raises the rollout size $K$ through 3, 5, and 8 for 20 steps each, with learning rate $2 \times 10^{-5}$. We keep the checkpoint with the highest mesh accuracy on validation problems, evaluated every 10 steps.
\section{Experimental setup}
\label{sec:setup}

\begin{figure*}[t]
  \centering
  \includegraphics[width=\textwidth]{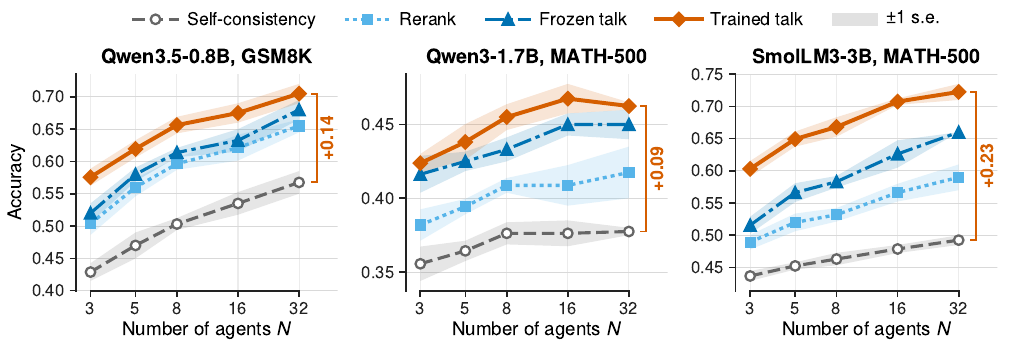}
  \caption{Accuracy against network size $N$ for Qwen3.5-0.8B on GSM8K and Qwen3-1.7B and SmolLM3-3B on MATH-500 (pooled evaluation on 200 test problems, Section~\ref{sec:setup}). Bands show one standard error. All conditions except SC use the centralized weighted vote. Labels give the margin of Trained talk over SC at $N{=}32$, rounded to two decimals.}
  \label{fig:scaling}
\end{figure*}

We use Qwen3.5-0.8B, Qwen3-1.7B, and SmolLM3-3B with thinking disabled and rank 16 LoRA adapters on all linear layers \citep{hu2022lora}. We evaluate the 0.8B model on GSM8K \citep{cobbe2021verifiers} and the larger models on the 326 numeric MATH-500 problems \citep{hendrycks2021math,lightman2023verify}. We use Qwen3.5-2B and Qwen3.5-4B for scaling within one model family (Section~\ref{sec:family}); Section~\ref{sec:attack} states its own test sets. Training uses the GSM8K training split and, for MATH-500, 4{,}958 numeric MATH training problems; training, validation, and test sets are disjoint.

We compare four conditions on identical proposals: (i) \emph{SC}, self-consistency by majority vote \citep{wang2023selfconsistency}; (ii) \emph{Rerank}, the weighted vote without communication; (iii) \emph{Frozen talk} and (iv) \emph{Trained talk}, the protocol with $M$ or $T$, respectively, writing hints and revisions. Reasoning agents sample at temperature 0.7 (nucleus 0.8, or 0.9 for SmolLM3-3B; top-$k$ 20); proposals and revisions use at most 320 (GSM8K) or 448 (MATH-500) new tokens.

Pooled evaluation uses 200 test problems per model (of 326 on MATH-500) and two pools of $P{=}32$ proposals per problem, scored once by $V$. Each $N \in \{3, 5, 8, 16, 32\}$ uses $\min(4, \lfloor P/N \rfloor)$ random subsets per pool, each rerunning stages (iii) and (iv). We average accuracy over subsets, then pools; standard errors combine both. Mesh accuracies use the centralized weighted vote. Gossip consensus ($R{=}25$ rounds, one graph per $N$) is compared with this vote and with SC (McNemar's exact test).
\section{Results}
\label{sec:results}

\subsection{Collective accuracy as a function of network size}
\label{sec:scaling}

We test whether Trained talk keeps its margin over SC as $N$ grows and how much of it communication causes. Figure~\ref{fig:scaling} reports all four conditions. At $N{=}32$, Trained talk exceeds SC by $+0.137$ (0.8B, GSM8K: $0.705$ against $0.568$), $+0.085$ (1.7B, MATH-500: $0.463$ against $0.378$), and $+0.230$ (3B, MATH-500: $0.722$ against $0.492$). With three agents, Trained talk reaches SC with 32 samples for every model ($0.576$ against $0.568$, $0.424$ against $0.378$, $0.603$ against $0.492$). Rerank reaches $0.655$, $0.417$, and $0.590$ at $N{=}32$, and Frozen talk $0.680$, $0.450$, and $0.660$. Trained talk exceeds both at every $N$, so training $T$ adds accuracy beyond the weighted vote and beyond hints and revisions written by $M$. Without the curriculum, a talk policy trained at fixed rollout size $K{=}5$ (not plotted) exceeds SC only for $N \leq 8$. Gossip consensus agrees with the centralized weighted vote on 99.4\% of problems averaged over models, $N$, and both pools (at least 98\% at every $N$). Its margin over SC is significant at every $N$ for all three models in both pools ($p<0.01$, McNemar's exact test).
\subsection{Scaling within one model family}
\label{sec:family}

\begin{table}[t]
  \centering
  \caption{GSM8K accuracy of Qwen3.5-0.8B, Qwen3.5-2B, and Qwen3.5-4B ( pooled evaluation on 200 test problems, two pools of 32 proposals; standard errors at most 0.020). Trained talk accuracy increases with model size at every $N$ (bold: highest accuracy at each $N$). Under self-consistency, the 2B model trails the 0.8B model at $N \geq 5$ (italics: largest gap).}
  \label{tab:family}
  \small
\begin{tabular}{l ccc ccc}
\toprule
 & \multicolumn{3}{c}{Self-consistency} & \multicolumn{3}{c}{Trained talk} \\
\cmidrule(lr){2-4} \cmidrule(lr){5-7}
$N$ & 0.8B & 2B & 4B & 0.8B & 2B & 4B \\
\midrule
3  & 0.429 & 0.432 & 0.579 & 0.576 & 0.708 & \textbf{0.726} \\
5  & 0.470 & 0.458 & 0.610 & 0.619 & 0.765 & \textbf{0.779} \\
8  & 0.503 & 0.473 & 0.621 & 0.656 & 0.804 & \textbf{0.809} \\
16 & 0.535 & 0.495 & 0.636 & 0.675 & 0.841 & \textbf{0.856} \\
32 & 0.568 & \textit{0.492} & 0.645 & 0.705 & 0.855 & \textbf{0.865} \\
\bottomrule
\end{tabular}
\end{table}

The models of Section~\ref{sec:scaling} differ in size and family. We therefore repeat confidence head training, talk policy training, and pooled evaluation on GSM8K for Qwen3.5-2B and Qwen3.5-4B. Table~\ref{tab:family} reports accuracy of SC and Trained talk. Under SC, the 2B model trails the 0.8B model at every $N \geq 5$. Trained talk reaches $0.705$, $0.855$, and $0.865$ for 0.8B, 2B, and 4B at $N{=}32$ and increases with model size at every $N$. For the 2B model at $N{=}32$, Trained talk exceeds SC by $0.362$ (from unrounded means over both pools), the largest margin over SC in this paper.
\subsection{Compromised agents and heterogeneous meshes}
\label{sec:attack}

\begin{figure*}[t]
  \centering
  \includegraphics[width=\textwidth]{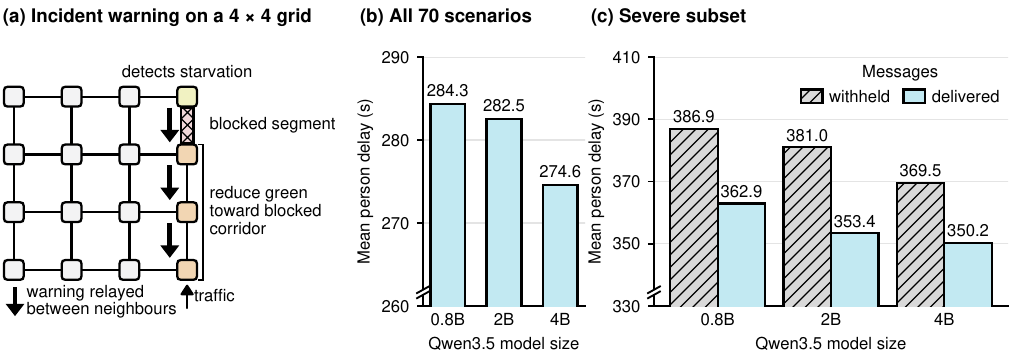}
  \caption{Traffic signal control. (a) The $4 \times 4$ grid with a blocked segment, an incident warning relayed between neighbouring intersections, and upstream metering. (b) Person delay of the Qwen3.5 meshes, all 70 scenarios, messages delivered. (c) The same meshes on the severe subset, messages withheld and delivered.}
  \label{fig:traffic}
\end{figure*}

In a coordinated attack, $k{=}4$ of $N{=}8$ agents submit a shared wrong answer with claimed confidence $0.99$ and, as speaker, send a hint asserting that answer. The defended mesh applies two checks: (i) agents rescore every solution with the shared confidence head, so all agents agree on the speaker and vote weights; (ii) revisions must pass the acceptance test. The undefended mesh omits both checks. With Trained talk on 150 test problems per model (0.8B on GSM8K; 1.7B and 3B on MATH-500), majority vote over all eight ballots falls to $0.000$ accuracy for every model and the undefended mesh to at most $0.007$. The defended mesh retains 0.507, 0.340, and 0.473 for 0.8B, 1.7B, and 3B (0.667, 0.513, 0.693 without attack) and exceeds the undefended mesh for every model ($p<10^{-3}$, McNemar's exact test). For 0.8B and 3B it exceeds SC over the four honest agents (0.473, 0.467).

Heterogeneous meshes on GSM8K ($N{=}8$) contain $(4,2,2)$, $(4,0,4)$, or $(0,4,4)$ agents of 0.8B, 2B, and 4B; each agent uses the confidence head and talk policy of its own model, without retraining. Over two runs per composition (same 200 test problems, independent proposals), the mesh averages 0.808 accuracy against 0.717 for SC; its lowest run (0.775) exceeds the highest SC run (0.740). Table~\ref{tab:instrumental} gives the range of mean accuracy per composition.
\subsection{Embodied coordination}
\label{sec:swarm}

On SwarmBench \citep{ruan2025swarmbench}, agents with $5{\times}5$ views on an $8{\times}8$ or $10{\times}10$ grid broadcast to visible agents. Flocking alone has dense reward, so REINFORCE trains moves and messages on the change in EMD score (reduction in earth mover's distance to the target shape) plus cohesion. With four agents, the EMD score over twenty unseen seeds rises from $0.05$ to $0.60$ (Qwen3.5-0.8B), $1.65$ to $2.30$ (Qwen3-1.7B), and $0.80$ to $0.95$ (SmolLM3-3B). The four other tasks count completions per episode and use distillation from a verifiable expert, a centralized solver: behaviour cloning on its traces, then GRPO on its scores of sampled moves, selecting checkpoints on separate seeds. On eighty unseen seeds, distilled Qwen3.5-0.8B matches the solver on Synchronization ($18/18$) and raises Foraging, Pursuit, and Transport from zero to 55\%, 48\%, and 26\% of the solver's $3.8$, $1.9$, and $5.2$.
\subsection{Traffic signal control under incidents and timing constraints}
\label{sec:traffic}

\begin{table*}[t]
  \centering
  \caption{Effect of withholding messages. Hidden regime: the acting agent cannot observe what its decision requires; observed regime: it can. $\Delta$: change from the control (accuracy; completions or captures per episode; person delay in s, negative is better). Ranges: compositions (two runs of 200 problems) or meshes (20 paired episodes). In parentheses: the largest reduction on one seed.}
  \label{tab:instrumental}
  \footnotesize
  \resizebox{\linewidth}{!}{\begin{tabular}{llccc}
\toprule
Setting & Control without communication & With communication & $\Delta$ & Regime \\
\midrule
Reasoning, 0.8B/GSM8K ($N{=}32$) & Rerank $0.655$ & Trained talk $0.705$ & $+0.050$ & hidden \\
Reasoning, 3B/MATH-500 ($N{=}32$)    & Rerank $0.590$ & Trained talk $0.722$ & $+0.132$ & hidden \\
Heterogeneous mesh ($N{=}8$, 2 runs each) & SC $0.705$ to $0.732$ & mesh $0.790$ to $0.830$ & $+0.085$ to $0.098$ & hidden \\
\midrule
SwarmBench Synchronization/Foraging/Pursuit & messages withheld & messages delivered & ${\approx}0$ ($<1$ s.e.) & observed \\
Pursuit, directive messages ($12{\times}12$, 8 agents) & withheld $1.16$ & delivered $1.12$ & ${\approx}0$ & observed \\
\midrule
Traffic, classical, single incident (40 seeds) & local adaptive $118.1$ & communicating $116.9$ & $-1.2$ ($-126$) & hidden \\
Traffic, classical, severe subset (10 scenarios) & local adaptive $379.3$ & communicating $331.2$ & $-48.2$ & hidden \\
Traffic, language model meshes, severe subset & withheld $370$ to $387$ & delivered $350$ to $363$ & $-19$ to $-28$ & hidden \\
\bottomrule
\end{tabular}}
\end{table*}

We simulate in SUMO \citep{lopez2018sumo} sixteen signalized intersections on a $4 \times 4$ grid of 200\,m blocks, modeled on the grid scenario of the RESCO benchmark \citep{ault2021resco}. Each intersection observes only its own camera. Timing constraints modeled on New York City signals fix a 90\,s cycle of two phases and a 12\,s minimum green, above the 7\,s minimum walk interval of the MUTCD \citep{fhwa2009mutcd}; the split (the division of green time between the phases) changes by at most $\pm 5$\,s per cycle. The objective is person delay: mean delay per person (s), counting 1.2 persons per car and one per pedestrian. In each of 40 incident seeds (one simulated hour), a random midblock avenue segment loses capacity for fifteen minutes. The downstream intersection observes the incident: the feeder from the blocked segment (an approach delivering vehicles into an intersection) stops delivering.

\textbf{Engineered controllers.} We compare (i) fixed time control, an even split that never adapts, (ii) a local adaptive controller that moves its split toward Webster's proportional split \citep{webster1958}, computed from the queues and waiting pedestrians it observes, and (iii) a communicating controller that adds messages to (ii). Each cycle, every intersection sends its neighbours its inbound queues, throughputs, and warnings. A starvation detector warns when a feeder with a running average of at least five vehicles per cycle delivers at most one while its queue is empty despite available green. Each intersection forwards a warning it receives to its neighbours; intersections on the flagged corridor apply metering, reducing their green toward it. Over these 40 seeds, mean person delay is $116.9$\,s (communicating), $118.1$\,s (local adaptive), and $157.0$\,s (fixed time). Against the local adaptive controller, the communicating controller has lower median ($75.7$ against $80.1$\,s) and maximum ($302.9$ against $342.8$\,s) person delay and reduces it by up to 126\,s on a single seed (Table~\ref{tab:instrumental}). The mean over the ten worst seeds ($\mathrm{CVaR}_{25\%}$) is $224.9$, $238.4$, and $309.4$\,s in the same order. A confirmatory sweep of 40 new seeds reproduces every ordering, including the lower maximum delay of the communicating controller (141\,s below local adaptive). An oracle controller metering from the true incident location and window has higher mean person delay ($124.5$\,s) than the communicating controller.

\textbf{Language model meshes.} We train Qwen3.5 meshes (0.8B, 2B, 4B) by behaviour cloning on traces of the communicating controller. Each prompt contains the camera summary, feeder deliveries against running averages, and received messages. Each response is a split change and a message of one line naming the nearest blocked corridor while the intersection holds a warning, or stating that none is blocked. We evaluate each mesh, controlling all sixteen intersections in closed loop, on 70 scenarios with three concurrent incidents of 25 minutes under two decoding strategies (greedy decoding and sampling at temperature 0.2). Every episode runs twice, with messages delivered and withheld. The severe subset is ten of these scenarios, fixed before any language model evaluation, on which the communicating controller reduces mean person delay by $48.2$\,s against the local adaptive controller (Table~\ref{tab:instrumental}); over all 70 scenarios, the reduction is $2.1$\,s. Training traces contain three concurrent incidents, and the training loss upweights responses with warnings.

\begin{table}[t]
  \centering
  \caption{Person delay (s) of the language model meshes. All scenarios: 140 paired episodes per model, messages delivered. Severe subset: 20 paired episodes per model.}
  \label{tab:traffic-llm}
  \small
\begin{tabular}{lccc}
\toprule
 & All scenarios & \multicolumn{2}{c}{Severe subset} \\
\cmidrule(lr){2-2} \cmidrule(lr){3-4}
Qwen3.5 mesh & Delivered & Withheld & Delivered \\
\midrule
0.8B & 284.3 & 386.9 & 362.9 \\
2B   & 282.5 & 381.0 & 353.4 \\
4B   & \textbf{274.6} & \textbf{369.5} & \textbf{350.2} \\
\bottomrule
\end{tabular}
\end{table}

Figure~\ref{fig:traffic} and Table~\ref{tab:traffic-llm} report the results. With messages delivered, mean person delay over all 70 scenarios is $284.3$, $282.5$, and $274.6$\,s for the 0.8B, 2B, and 4B meshes. On the severe subset, delivering messages reduces the person delay of every mesh; the pooled reduction over 60 paired episodes is $23.6 \pm 6.7$\,s (mean and standard error). Over all 70 scenarios, messages reduce the person delay of the 4B mesh by $11.4 \pm 6.9$\,s. For the 0.8B and 2B meshes, the change matches the $2.1$\,s reduction of the communicating controller within one standard error. Scored against the true incidents, warnings of the three meshes have mean precision 0.76 (share of warnings naming a blocked corridor during an incident or the six cycles after) and recall 0.98 (share of incident cycles with a warning on a blocked corridor within two cycles). Corridor corroboration, unused in Table~\ref{tab:traffic-llm}, delivers a warning only after two distinct senders flag one corridor within two cycles. Under corroboration, one compromised intersection that keeps its control actions but warns of its own unblocked corridor every cycle leaves every mesh's trajectory unchanged (severe subset, greedy decoding).
\subsection{An observability criterion for communication}
\label{sec:instrumental}

Controls remove messages without retraining, in three hidden and two observed settings (Table~\ref{tab:instrumental}). In reasoning, listeners see the speaker's solution only in the hint: Rerank (no hint, no revision) lowers accuracy at $N{=}32$ by 0.050, 0.046, and 0.132 (0.8B on GSM8K; 1.7B and 3B on MATH-500). In heterogeneous meshes, hints cross model sizes; their control, SC, also omits confidence weighting. Only the downstream intersection observes a traffic incident; upstream intersections act on it. On the severe subset, messages reduce person delay for every language model mesh. On Synchronization, Foraging, and Pursuit, each agent's $5{\times}5$ view shows what its decision requires; Synchronization stays at its ceiling without messages, and Foraging and Pursuit scores with and without messages agree within one standard error. Directive messages that assign moves to agents (Pursuit, $12{\times}12$ grid) yield 1.12 captures per episode, against 1.16 without. All outcomes match the criterion.
\section{Limitations and future work}
\label{sec:limitations}

All evaluated tasks have automatically scored outcomes. Training requires supervision: (i) ground truth answers in reasoning, (ii) task rewards and a centralized solver on SwarmBench, and (iii) traces of the communicating controller in traffic. The method therefore needs labelled problems near the target distribution. Traffic results come from SUMO simulation under New York City timing constraints. The scaling study covers one family, Qwen3.5 from 0.8B to 4B, our largest model. Future work will extend the acceptance test and gossip consensus to (i) agents with separate owners and incentives and (ii) networks with partitions, churn, and asynchrony.

\paragraph{Reproducibility.} We release all environments, training scripts, evaluation protocols, seeds, result files, and the exact command for every reported number.
\section{Conclusion}
\label{sec:conclusion}

With three agents, \sysname{} reaches the accuracy of SC with 32 samples; at $N{=}32$, it raises accuracy from $0.492$ to $0.722$ (SmolLM3-3B, MATH-500) and from $0.568$ to $0.705$ (Qwen3.5-0.8B, GSM8K), and communication and training each add accuracy beyond the weighted vote. With four of eight agents compromised, majority vote accuracy falls to $0.000$, while the defended mesh retains $0.507$ (Qwen3.5-0.8B, GSM8K) and significantly exceeds the undefended mesh for every model. Over 40 incident seeds, the communicating traffic controller has the lowest mean person delay ($116.9$\,s), and on the severe subset messages reduce the person delay of every Qwen3.5 mesh ($23.6$\,s pooled over 60 paired episodes). Across all settings, messages improve a decision when another agent holds the information it requires.

\bibliographystyle{abbrvnat}
\bibliography{references}
\end{document}